\documentclass[lettersize,journal]{IEEEtran}
\usepackage{amsmath,amsfonts}
\usepackage{algorithmic}
\usepackage{algorithm}
\usepackage{array}
\usepackage[caption=false,font=normalsize,labelfont=sf,textfont=sf]{subfig}
\usepackage{textcomp}
\usepackage{stfloats}
\usepackage{url}
\usepackage{verbatim}
\usepackage{graphicx}
\usepackage{cite}
\usepackage{microtype}      
\usepackage{xcolor}         

\usepackage{times}
\usepackage{epsfig}
\usepackage{amsmath}
\usepackage{amssymb}
\usepackage{enumitem}
\usepackage{float}
\usepackage{booktabs}
\usepackage{color}
\usepackage{soul}
\usepackage{colortbl}
\usepackage{multirow}
\usepackage{caption}
\newcommand{\myparagraph}[1]{\smallskip \noindent{\bf {#1}.}}

\usepackage{natbib}
\small
\newcommand{\projecttitle}{MultiGO++}

\newcommand{\red}[1]{\textcolor{black}{#1}}

\begin{document}
\title{MultiGO++: Monocular 3D Clothed Human Reconstruction via Geometry-Texture Collaboration} 
\author{Nanjie Yao*, Gangjian Zhang*, Wenhao Shen, Jian Shu, Yu Feng, and Hao Wang$^\dagger$}

\twocolumn[{
\renewcommand\twocolumn[1][]{#1}
\begin{center}
    \maketitle
    \centering
    \captionsetup{type=figure}
    \includegraphics[width=1\textwidth]{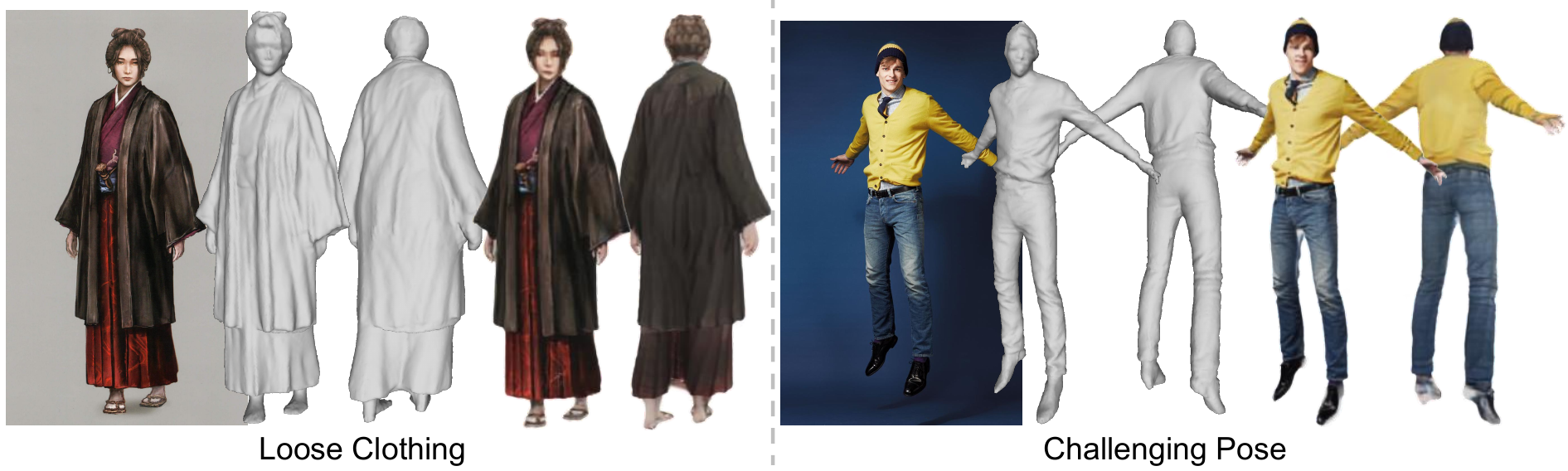}
    \vspace{-0.7cm}
    \captionof{figure}{\textbf{Monocular 3D human reconstruction on challenge in-the-wild cases.} The proposed {\projecttitle}  exhibits strong generalization and robustness, even in these difficult in-the-wild cases, such as those shown in the figure.}
    \label{fig: head}
\end{center}
}]

\renewcommand{\thefootnote}{}
\footnotetext{\noindent Nanjie Yao, Gangjian Zhang, Jian Shu, Yu Feng, and Hao Wang are with The Hong Kong University of Science and Technology (Guangzhou), Guangzhou 511442, China (e-mail: nanjieyao@gmail.com, gzhang292@connect.hkust-gz.edu.cn, haowang@hkust-gz.edu.cn).}
\footnotetext{\noindent Wenhao Shen is with Nanyang Technological University, Singapore 639798, (e-mail: wenhao005@e.ntu.edu.sg)}
\footnotetext{\noindent $\dagger$: Corresponding author, *: Equal contribution} 


\begin{abstract}
Monocular 3D clothed human reconstruction aims to generate a complete and realistic textured 3D avatar from a single image. Existing methods are commonly trained under multi-view supervision with annotated geometric priors, and during inference, these priors are estimated by the pre-trained network from the monocular input.
These methods are constrained by three key limitations: texturally by unavailability of training data, geometrically by inaccurate external priors, and systematically by biased single-modality supervision, all leading to suboptimal reconstruction. 
To address these issues, we propose a novel reconstruction framework, named MultiGO++, \red{which achieves effective systematic geometry-texture collaboration.} \red{It} consists of three core parts: (1) A multi-source texture synthesis strategy that constructs 15,000+ 3D textured human scans to improve the performance on texture quality estimation in challenge scenarios; (2) A region-aware shape extraction module that extracts and interacts features of each body region to obtain geometry information and a Fourier geometry encoder that mitigates the modality gap to achieve effective geometry learning; (3) A dual reconstruction U-Net that leverages \red{geometry-texture} collaborative features to refine and generate high-fidelity textured 3D human meshes. Extensive experiments on two benchmarks and many in-the-wild cases show the superiority of our method over state-of-the-art approaches. Our project page can be seen at: \url{https://3dagentworld.github.io/multigo++}.




\end{abstract}


\section{Introduction}~\label{sec: intro}
Creating a photorealistic, full-body, and clothed 3D human avatar from a single image is crucial for numerous industries, such as gaming, film, augmented reality, virtual reality~\cite{10669838, 10836818, 10814680, 10168294}. This process involves generating a complete 3D human avatar of a person solely based on a single RGB image. 
However, given that the input image only provides a front view, the missing texture information in invisible regions and the ambiguity in geometric estimation hinder the reconstruction of a photorealistic 3D human avatar.

To mitigate this problem, existing methods~\cite{zheng2020pamir, li2022neurips_fof, zhang2024global_gta, Zhang_2024_sifu, hilo, VS_CVPR2024, xiu2023econ, xiu2022icon, zhang2024multigo, li2024pshuman, 10423797}, such as SiTH~\cite{ho2024sith}, typically introduce explicit external priors, represented by SMPL-related body mesh and synthetic images. Specifically, these approaches use rendered monocular images and corresponding annotated explicit external priors from 3D human scan datasets to train the models. During inference, given a monocular image, they first employ a human pose and shape estimation model~\cite{loper2023smpl, pavlakos2019expressive_smplx, romero2022embodied_smplh} or a novel view synthesis model~\cite{wang2023imagedream} to estimate the required priors. They then combine these priors with the input image and feed them into a subsequent 3D human reconstruction model for avatar modeling and reconstruction.

However, such methods still face the following limitations: (1) from the texture perspective, the scarcity of 3D human scans significantly limits the quality of reconstructed textures and their generalization in complex scenarios; 
(2) from the geometric perspective, inaccurate explicit external priors used in the inference stage inevitably weaken the accuracy of the reconstructed geometry~\cite{li2024pshuman, hilo};
and (3) from the systematic perspective, existing methods~\cite{zhang2024multigo, 3DGaussian} only use multi-view images as texture training supervision, which causes the model to often ignore the output geometric accuracy.

To address the above issues, we design a new collaborative monocular human reconstruction framework, named \projecttitle. It comprises three major parts: (1) we propose 
a multi-source texture synthesis strategy, leveraging existing text-to-3D~\cite{huang2023humannorm, star} and image-to-3D~\cite{li2024pshuman} models to generate diverse synthetic textured 3D human as the training data. We also employ a multimodal large language model (LLM)~\cite{openai2024gpt4ocard} to ensure generation quality. 
We construct a synthetic dataset of over 15,000 high-quality 3D human scans to improve our performance in texture prediction; (2) in the geometry part, we design a cross-attention-based region-aware shape extraction module 
to extract the features of segmented body regions from the input monocular image to obtain relevant human shape information.
Then we utilize Fourier expansion, interpolation, and projection to bridge the modality gap between 2D texture and 3D geometry, such that the output geometry can be enhanced; and (3) we propose a dual reconstruction U-Net, 
consisting of one normal Gaussian avatar U-Net and one textured Gaussian avatar U-Net. 
Furthermore, a Gaussian-enhanced remeshing strategy is proposed to efficiently generate human meshes by leveraging the normal Gaussian avatars.


Extensive experiments show that the proposed method surpasses existing state-of-the-art~(SOTA) monocular human reconstruction approaches. Additionally, more in-the-wild cases further confirm the generalization and practicality of our proposed method. The key contributions of this paper can be summarized as: 
\begin{itemize} 

\item Texturally, we design a multi-source texture synthesis strategy that aggregates off-the-shelf $\mathbf{X}$-to-3D models from various 3D domains to construct synthetic 3D human scan training data with different appearances. These data further enhance the texture prediction performance, particularly for challenging in-the-wild cases. 

\item Geometrically, we construct a region-aware shape extraction module that achieves effective 3D human shape feature extraction and a Fourier geometry encoder that integrates 2D texture and 3D geometry features. These modules improve the error propagation during inference and bridge the gap between cross-modal features to achieve efficient and robust monocular 3D human geometry feature extraction.

\item Systematically, we propose a dual reconstruction U-Net which integrates and interacts geometric and textural features and utilizes the cross-modal output to achieve post-processing optimization of the coarse human mesh for high-quality texture prediction and lossless human mesh reconstruction. 

\end{itemize}

Our preliminary research has been published in~\cite{zhang2024multigo}. The code is publicly available$^\S$~\footnote{$\S$ Preliminary research code: \url{https://github.com/gzhang292/MultiGO}}.

\section{Related Work} \label{sec: relatedwork}
\myparagraph{Single-view 3D Human Reconstruction} 
\red{Reconstructing and understanding 3D representations from 2D inputs is a fundamental challenge in computer vision~\cite{drawing2cad, vgnet, fusgcn,tang2024lgm,hong2023lrm,zhang2024gslrm, cheng2025unposed, shu2025fastanimate}.} Reconstructing 3D human models from monocular input has garnered more attention in recent research~\cite{shen2025smpl, zhang2025sat}. The first approach, PIFu~\cite{saito2019pifu}, introduces a pixel-aligned implicit function that enables shape and texture generation. Following this approach, many methods, represented by ICON~\cite{xiu2022icon}, improve the quality of reconstruction by introducing parametric models such as SMPL~\cite{loper2023smpl} and SMPL-X~\cite{pavlakos2019expressive_smplx} as human body prior. Building on ICON, ECON~\cite{xiu2023econ} enhanced the method with explicit body regularization. Subsequently, GTA~\cite{zhang2024global_gta} leverages transformer architectures to capture global-correlated image features, and HiLo~\cite{hilo} introduces an approach leveraging high and low frequency features. In achieving real-time inference, FOF~\cite{li2022neurips_fof} proposes an efficient 3D representation by learning the Fourier series and its extension FOF-X~\cite{feng2024fof}, which avoids the performance degradation caused by texture and lighting. R$^2$Human~\cite{yang2024r} introduces a novel representation to achieve real-time rendering. In addressing challenges related to loose clothing, VS~\cite{VS_CVPR2024} proposes a stretch-based method to improve reconstruction quality. Current methods improve the reconstruction quality by introducing diffusion models. SiTH~\cite{ho2024sith} utilizes a 2D diffusion model to enhance occlusion area predictions. HumanRef~\cite{zhang2024humanref} employs an optimization approach with the proposed reference-guided score distillation to generate a textured 3D human avatar. PSHuman~\cite{li2024pshuman} designs a global-local diffusion backbone and introduces a noise blending mechanism during diffusion denoising to improve the quality of facial reconstruction. 

\myparagraph{Gaussian Model for Human Reconstruction} Recent advancements in 3D human digitalization have explored the use of Gaussian Splatting~\cite{kerbl3Dgaussians} as a novel 3D representation. For video-based inputs, Gauhuman~\cite{hu2023gauhuman} proposes optimization-based approaches to refine the human Gaussians. When dealing with sparse-view inputs, GPS-Gaussian~\cite{GPS-Gaussian} and EVA-Gaussian~\cite{evagaussian} introduce a generalizable multi-view framework for reconstructing high-fidelity human Gaussian avatars. For single-view inputs, MultiGO~\cite{zhang2024multigo} presents a multi-level reconstruction framework that tackles the challenges of limited training data. Human3Diffusion~\cite{3diffusion} integrates a 2D multi-view diffusion model into a 3D reconstruction framework and designs a 2D-3D joint training paradigm to enhance 3D Gaussian generation. HGM~\cite{hgm} adopts a generate-then-refine pipeline, achieving improved performance on texture estimation for invisible parts.

For the monocular input setting, while these methods have made significant strides, challenges remain, including addressing inaccuracies of estimated geometry prior in the inference stage and mitigating the scarcity of training data to improve the model's generalization ability. Existing approaches still lack effective solutions to these issues, leading to suboptimal reconstruction quality.

\myparagraph{Human Pose and Shape Estimation} 
In the domain of Human Pose and Shape (HPS) estimation from monocular images, the goal is to reconstruct a 3D human body mesh, typically parameterized using models such as SMPL~\cite{loper2023smpl}, SMPL-X~\cite{pavlakos2019expressive_smplx}, and SMPL-H~\cite{romero2022embodied_smplh}. Early works in this area predominantly adopt optimization-based strategies~\cite{kolotouros2019learning}. These methods iteratively fit a parametric model to 2D observations—such as keypoints~\cite{smplify} by minimizing an objective function composed of data terms (measuring reprojection errors) and prior terms (penalizing implausible poses or shapes). Subsequent improvements integrate richer cues into the optimization process, including 2D/3D joints, segmentations, and dense correspondences. In contrast to optimization-based techniques, regression-based methods harness the powerful nonlinear mapping capabilities of deep neural networks to directly predict parametric model coefficients from raw image pixels~\red{\cite{zhang2021pymaf, zhang2023pymaf, pixie, osx, hmradapter, Goel_2023_ICCV}}. This paradigm shift enables single-shot inference, bypassing the iterative fitting process and its associated computational cost. A significant body of research has focused on designing novel network architectures and regression targets to improve accuracy and robustness. 

In monocular 3D human reconstruction, approaches such as PyMAF~\cite{zhang2021pymaf}, PyMAF-X~\cite{zhang2023pymaf}, SMPLify-X~\cite{smplify}, and PIXIE~\cite{pixie} are commonly employed to predict SMPL-related parameters in inference. However, they are fundamentally constrained by the inherent ambiguity in a single input view, often resulting in dissatisfactory depth estimation and eventually leading to reduced reconstruction accuracy in the inference stage.

\begin{figure*}
    \centering
    \includegraphics[width=0.98\linewidth]{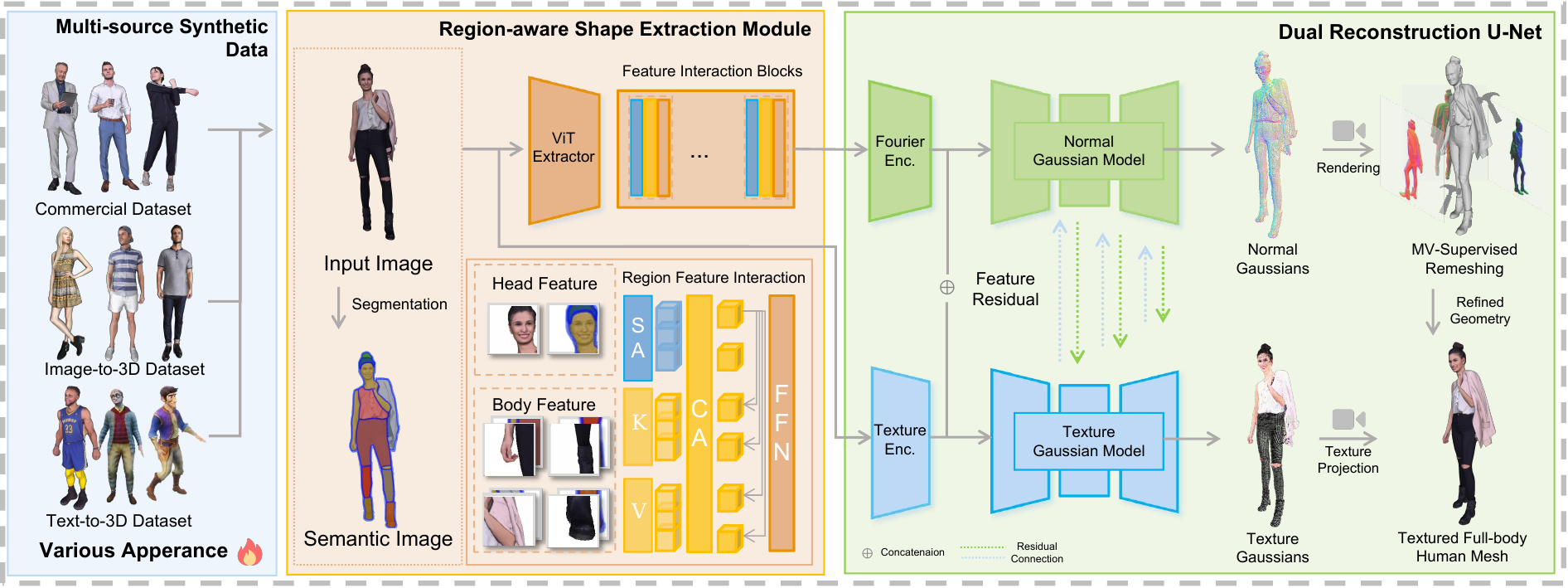}
    \vspace{-0.1cm}
    \caption{\textbf{Method Overview.} Our framework integrates three core components: Texturally, we employ a multi-source texture synthesis strategy to generate diverse synthetic data for training, along with a lightweight texture encoder for effective feature extraction. \red{Geometrically, we introduce a Region-aware Shape Extraction Module that enhances human shape extraction through part-based feature interaction, utilizing Self-Attention (SA), Cross-Attention (CA), and Feed-Forward Networks (FFN). This is coupled with a Fourier Geometry Encoder to bridge the modality gap for efficient geometric learning. }Systematically, we propose a Dual Reconstruction U-Net that utilizes feature residuals to balance geometric and texture features, enabling mutual enhancement across modalities. Additionally, to refine 3D mesh quality and extraction efficiency, we design a Gaussian-enhanced remeshing strategy supervised by the generated normal Gaussian avatar.}
    \label{fig: overview}
    \vspace{-0.5cm}
\end{figure*}

\section{Methodology}\label{sec: method}
\subsection{Preliminaries}\label{sec: prelim}
\myparagraph{Gaussian Splatting} Gaussian Splatting, introduced by Bernhard et al.~\cite{kerbl3Dgaussians}, represents a 3D scene or asset using a collection of 3D Gaussians. Each Gaussian is defined by a set of attributes: a geometric center $x \in \mathbb{R}^3$, a scaling factor $s \in \mathbb{R}^3$, a rotation quaternion $r \in \mathbb{R}^4$, an opacity $\alpha \in \mathbb{R}$, and a color descriptor $c \in \mathbb{R}^3$. Together, a 3D asset is explicitly represented as a set of Gaussians $G = \{G_i\}$, where each 3D Gaussian $G_i = \{x_i, s_i, r_i, \alpha_i, c_i\} \in \mathbb{R}^{14}$ encapsulates the attributes of the $i$-th component.

\myparagraph{SMPL-X Model} The Skinned Multi-Person Linear (SMPL) model~\cite{loper2023smpl} is widely used in the fields of Human Pose and Shape (HPS) estimation and 3D human reconstruction. We build our \projecttitle~on one of its variant models, SMPL-X~\cite{pavlakos2019expressive_smplx}. SMPL-X utilizes a set of input parameters: body pose (including global orientation, hands and jaw poses) $\theta \in \mathbb{R}^{53 \times 3}$, expressed in the axis-angle representation; body shape $\beta \in \mathbb{R}^{10}$; and facial expression $\alpha \in \mathbb{R}^{10}$. These parameters define a human body mesh $\mathbf{M}$ as follows: $\mathbf{M} = \text{SMPL-X}(\theta, \alpha, \beta) \in \mathbb{R}^{\mathcal{V} \times 3}$, where $\mathcal{V} = 10,475$ represents the number of vertices.

\subsection{Texture: Multi-source Texture Synthesis Strategy}
\label{sec: texture}

\begin{figure}
    \centering
    \includegraphics[width=1\linewidth]{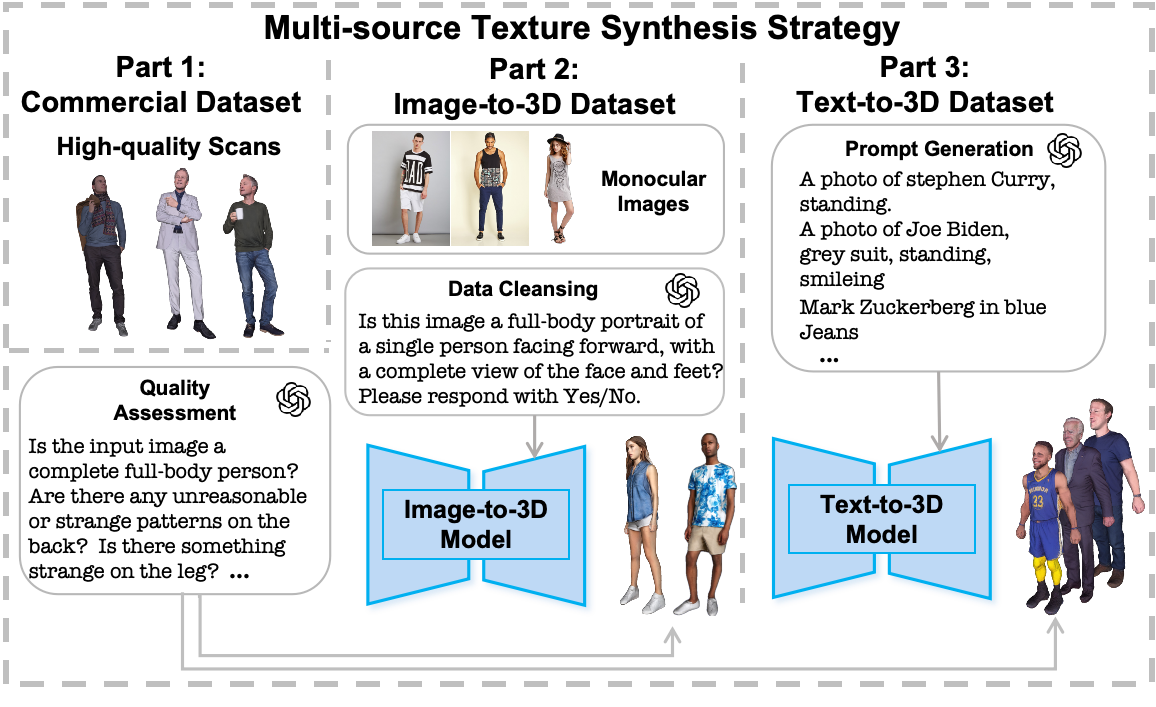}
    \vspace{-0.7cm}
    \caption{\textbf{Multi-source Texture Synthesis Strategy.} The proposed multi-source texture synthesis strategy leverages X-to-3D models and multimodal LLM data screening to generate high-quality training data for enhanced texture estimation.} 
    \label{fig: aug}
    \vspace{-0.4cm}
\end{figure}

\myparagraph{Synthetic Texture} As discussed in Sec.~\ref{sec: intro}, from the texture perspective, existing methods are largely constrained by the scarcity of 3D human scan data for training, leading to suboptimal performance on challenging inputs. To address this limitation and boost our model’s performance and generalization—especially for out-of-distribution and in-the-wild challenging inputs—we propose an innovative multi-source texture synthesis strategy. This strategy aims to construct a training dataset with diverse textured appearances, containing over 15K samples. Beyond open-source datasets, our data sources include commercial datasets, along with image-to-3D and text-to-3D generated data. The dataset structure is detailed as follows:


\textbf{1)} For commercial data, we collect ~3K high-quality 3D human scans from publicly accessible commercial repositories~\cite{renderpeople, axyz, twindow, treedy}.
\textbf{2)} For image-to-3D generated data, we first gather over 200,000 real-world images from relevant datasets~\cite{liu2016deepfashion, cosmicman}. A multimodal LLM~\cite{openai2024gpt4ocard} is used for initial data screening and cleaning, yielding 50,000 high-quality, full-body photorealistic human images (see Part 2 of Fig.~\ref{fig: aug}). These images are then input to diffusion-based image-to-3D synthesis models~\cite{li2024pshuman, huang2024tech} to generate additional high-fidelity synthetic 3D human scans. To ensure quality and reduce hallucinations in occluded areas, a second multimodal LLM-based quality assessment is conducted, ultimately retaining over 10,000 high-quality samples.
\textbf{3)} For text-to-3D generated data (see Part 3 of Fig.~\ref{fig: aug}), an LLM is used to automatically generate over 5,000 prompts describing humans with diverse clothing, appearances, and poses. These prompts are fed into text-to-3D models~\cite{huang2023humannorm, star} to synthesize various human scans. Consistent with the image-to-3D pipeline, LLM-based quality assessment is performed, resulting in ~1,000 high-quality samples.


To sum up, our dataset comprises over 15,000 high-quality 3D human scans, covering a wide range of appearances, poses, and clothing. 

\myparagraph{Texture Encoder} To enable efficient texture feature extraction while preserving spatial dimension alignment with our geometry representation, we adopt a lightweight texture encoder. Specifically, this encoder consists of a single convolutional layer followed by a spatial attention module. For the frontal input image (denoted as $\mathcal{I}_{0} \in \mathbb{R}^{3 \times H \times W}$), we first concatenate it along the channel dimension with a corresponding Plücker ray camera feature (which encodes camera poses). This concatenated input is then fed into the texture encoder to extract texture features, represented as $\mathcal{F}_{c} \in \mathbb{R}^{1 \times o \times H \times W}$, where $o$ is the number of channels, and $H$, $W$ (the height and width of the output feature map) match those of the input image.



\subsection{Geometry: Shape Extraction \& Geometry Learning}
\label{sec: geometry} 

\myparagraph{Region-aware Shape Extraction Module} As analyzed in Sec.~\ref{sec: intro}, the monocular setting of this task implies that frontal human RGB images alone cannot provide sufficient geometric information. While traditional HPS estimation models are introduced to address this issue, they inevitably degrade the reconstruction model’s performance—this is due to the inaccurate geometric representations estimated during inference.
To tackle this problem, we propose a region-aware shape extraction module, which extracts human shape-related features from the monocular input image. This module replaces the conventional, widely used HPS estimation pipeline. Furthermore, it eliminates reliance on annotated geometric priors, allowing the model to scale more effectively. It also indirectly fulfills the training augmentation objective proposed in previous work~\cite{zhang2024multigo}, thereby improving the qualitative robustness of the reconstruction model. The detail of this module is illustrated in the middle part of Fig.~\ref{fig: overview}.


Given the input image, we first leverage a pre-trained semantic segmentation network~\cite{kirillov2023segment} to obtain semantic masks corresponding to various parts of the human body, denoted as $\mathcal{S} = \{ {s}^{i}| i=0,...,k\}$. Here, $i$ represents the ordinal number of different semantic masks, which include the head, torso, hands, lower limbs, arms, and more. We then crop distinct regions to create square rectangles using the mask boundary coordinates and resize them to the same size. This process yields a set of subgraphs, denoted as $\mathcal{G} = \{ {g}^{j} \in \mathbb{R}^{3\times H\times W}| j=0,...,m\}$, where $m$ is the number of subgraphs. These subgraphs are individually processed into features using a pretrained vision transformer~\cite{vitpose, dosovitskiy2020image} to produce body local features $\textbf{T}_{body}$.


To facilitate comprehensive information exchange across the human body within each patch, we design a feature interaction block based on a cross-attention architecture~\cite{crossattention}. Specifically, we utilize the head feature $\textbf{T}_{head}$ as a primary keypoint~\cite{duan2019centernet} to determine the human position in the input image. We then treat $\textbf{T}_{head}$ as an initialized cross-attention query $Q$, while the body features $\textbf{T}_{body}$ serve as both keys and values, represented as $K$ and $V$. The query is updated through self-attention layers ($\textbf{SAttn}$), a cross-attention layer ($\textbf{CAttn}$), and a Multi-Layer Perceptron ($\textbf{MLP}$). This attention mechanism allows the query features, akin to anchor features, to effectively absorb depth information from various levels across the body. This process can be expressed as: 
\begin{align} 
Q' = \textbf{MLP}(\textbf{CAttn}(\textbf{SAttn}(Q),\ [K,V])). 
\end{align}

The updated query $Q'$ from the feature interaction block is subsequently transformed into a human body mesh as a geometric representation through MLP layers and an SMPL-X layer, denoted as $p$.

\begin{figure}
    \centering
    \includegraphics[width=1\linewidth]{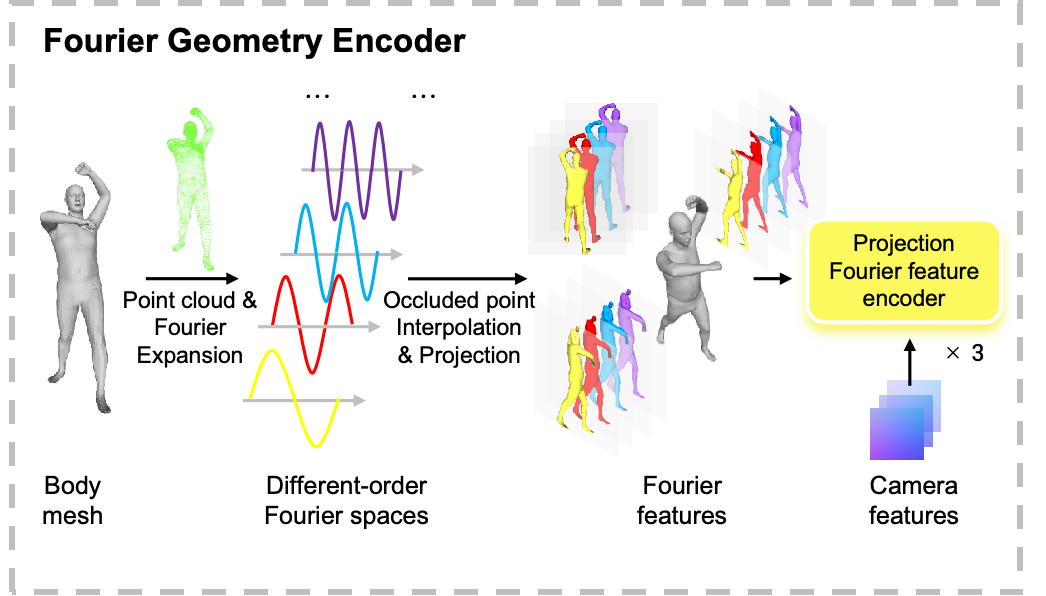}
    \caption{\textbf{Detailed Architecture of Fourier Geometry Encoder.} To achieve effective geometry learning, we achieve better fusion of the heterogeneous modalities of the 3D geometry prior and 2D images. We propose interpolating the Fourier features of 3D occluded points and mapping them from three different angles into the same 2D space as the image features. } 
    \label{fig: fourier}
    \vspace{-0.5cm}
\end{figure}

\myparagraph{Fourier Geometry Encoder} Through the proposed region-aware shape extraction module, we obtain a human body mesh that captures human geometry. Recognizing that texture and geometric features stem from two distinct modalities with a large semantic gap, our approach avoids rigid fusion of these cross-modal features. Instead, the Fourier geometry encoder further projects 3D Fourier features into the same 2D space as the input image features, enabling better interaction and fusion of these heterogeneous features. This module allows the model to effectively learn human geometry. The detailed architecture of the Fourier geometry encoder is shown in Fig.~\ref{fig: fourier}.

Concretely, inspired by some works~\cite{li2024craftsman,3DShape2VecSet}, the proposed Fourier geometry encoder first considers all vertices of the given $p \in F_{k}$ as points of the point cloud. The point cloud can be represented as $p \in \mathbb{R}^{3\times10475}$. Then, the 3D Fourier expansion operation is used to enhance the expression of these points. Specifically, we extract $q$-order Fourier series for each point $p$ in $F_{k}$ as follows:
\begin{equation} \label{mand1-m}
    \mathcal{F}(p)  = \left\{p\right\} \cup \left\{\cos(2^{n}p), \sin(2^{n}p)|n \in\left\{1,...,q\right\}\right\}.
\end{equation}
Through the above operation, we have expanded the 3D space where the given points of geometric feature are located into the different Fourier spaces $\left\{\mathcal{S}_{n} | n \in\left\{0,...,2q\right\} \right\}$. The point clouds in these spaces are denoted as $\left\{\tilde{\mathcal{P}_{n}} | n \in\left\{0,...,2q\right\} \right\}$. Meanwhile, we interpolate and expand them to make the point clouds in these spaces denser. Specifically, we interpolate positions on the surface of a triangular surface and average the weights of three points belonging to the same triangular surface. After this, denser point clouds $\tilde{\mathcal{P}_{n}} \in \mathbb{R}^{3\times m}$ with different-order Fourier are obtained, where $m$ is the point number.

To facilitate the fusion of geometric and texture features, we perform 2D projection on the occluded points in different Fourier spaces from three camera angles. By doing so, we can obtain a stack of Fourier features from different spaces, which can be concatenated into $\tilde{\mathcal{F}_{1}} \in \mathbb{R}^{3(2q+1) \times H \times W}$, where $H$ and $W$ are the resolution of the projection plane. Similarly, from the perspectives of the other two cameras, we can obtain $\tilde{\mathcal{F}_{2}}$ and $\tilde{\mathcal{F}_{3}}$. Subsequently, all of them, along with their camera feature, are fed into a Fourier feature encoder to obtain geometric features, $\mathcal{F}_{1}^{\prime}, \mathcal{F}_{2}^{\prime}, \mathcal{F}_{3}^{\prime} \in \mathbb{R}^{o \times H \times W}$. The Fourier feature encoder consists of a single 2D convolutional layer, configured with a kernel size of 3, a stride of 1, and padding of 1. This configuration is chosen to preserve the spatial dimensions of the feature map, thereby aligning its output dimensionality with that of the reconstruction backbone's input. The encoder outputs are then concatenated into the Fourier geometric feature, denoted as $\mathcal{F}^g \in \mathbb{R}^{3 \times o \times H \times W}$.

\begin{figure*}[t!]
    \centering
    \includegraphics[width=1\linewidth]{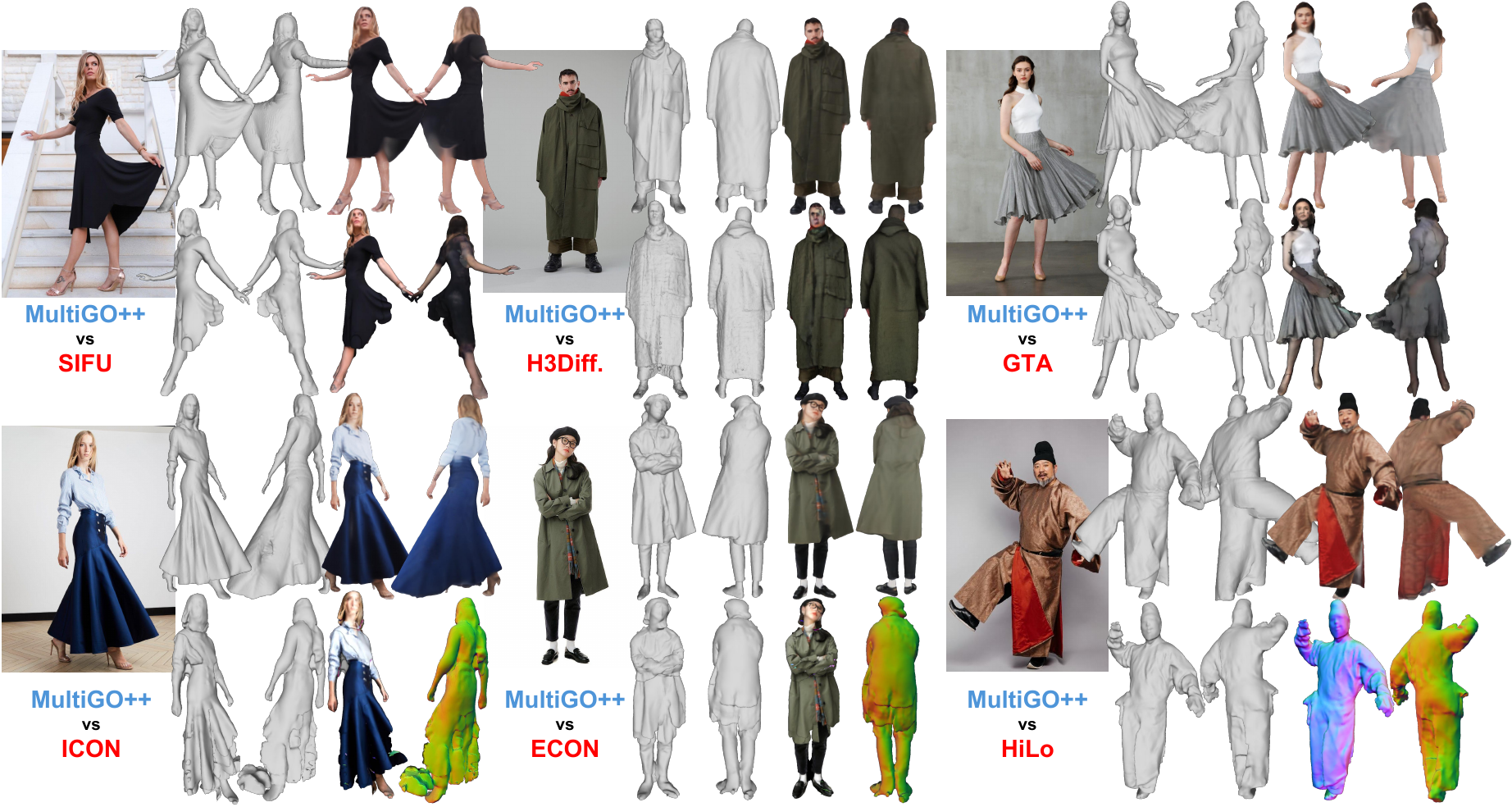}
    \vspace{-0.5cm}
    \caption{\red{\textbf{Qualitative comparisons on in-the-wild images featuring loose clothing.} While other SOTA methods struggle to accurately reconstruct the challenging geometries of loose garments, our approach faithfully reproduces high-fidelity wrinkles and intricate textures.} Please \textbf{zoom in}~\includegraphics[width=0.015\textwidth]{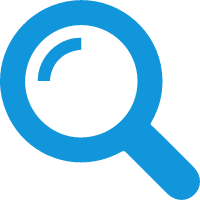}~for a detailed view.}
    \label{fig: viswild}
    \vspace{-0.5cm}
\end{figure*}

    

\subsection{System: Dual Reconstruction U-Net}


\myparagraph{Biased Feature Learning} As depicted in Sec.~\ref{sec: geometry} and Sec.~\ref{sec: texture}, we extract texture features $\mathcal{F}_{c}$ from the texture module and Fourier geometric features $\mathcal{F}_{g}$ from the geometry module, respectively. This setup enables bidirectional information transfer—allowing texture details to inform geometric representations and vice versa. However, since our training for textured Gaussian avatar prediction relies on 2D RGB data as supervision (following~\cite{zhang2024multigo}), this inherent imbalance prioritizes texture feature learning, diminishing the model’s focus on geometric features. To address this bias, we propose a dual reconstruction U-Net, specifically designed to enhance attention to geometric aspects.

Aligned with prior work~\cite{zhang2024multigo}, the dual reconstruction U-Net first concatenates $\mathcal{F}_{g}$ and $\mathcal{F}_{c}$ to form a combined feature representation. This fused feature is then fed into a pre-trained U-Net to predict a 3D textured Gaussian avatar. For supervision, we render RGB images from both the predicted textured Gaussian avatar and ground-truth 3D scans—using an identical camera system for consistency—and minimize discrepancies between these renderings via 2D losses (MSE loss, mask loss, and LPIPS loss). Notably, this texture-focused supervision still tends to overshadow geometric information extraction. To counterbalance this, we design a parallel U-Net branch dedicated to normal Gaussian avatar prediction:
\begin{equation}
G^{c} = R_c(\mathcal{F}_{g}, \mathcal{F}_{c});\ \ G^{n} = R_n(\mathcal{F}_{g}, \mathcal{F}_{c}),
\end{equation}
where $R_c$ and $R_n$ are the texture reconstruction network and normal reconstruction network. $G^{c}$ and $G^{n}$ represent the predicted textured Gaussians and normal Gaussians, respectively\footnote{To clarify, the "normal Gaussian" herein does not refer to the normal vector of 3D Gaussian Splatting (3DGS); instead, it refers to the 3DGS used to construct the normal avatar.}. To strengthen the learning connection between these two reconstruction networks—enabling them to mutually reinforce each other—we propose a feature exchange mechanism based on cross-U-Net residuals. In detail, we decompose each U-Net into three distinct stages: the Encoder (Down blocks), Bottleneck (Middle block), and Decoder (Up blocks).

During the forward pass, features are initially in parallel processed by the Down-Blocks of both U-Nets. This process utilizes the inherent encoder-decoder architecture, allowing each modality-specific network to extract relevant features through its respective encoder. The encoded features are then passed to the Mid-Blocks, resulting in feature maps $\mathcal{F}{c_0}$ and $\mathcal{F}{n_0}$ from the Mid-Blocks of the two U-Nets, denoted as $MB_c(\cdot)$ and $MB_n(\cdot)$.

To integrate these features, we employ a linear residual connection, producing a fused feature map: $\mathcal{F}{f_0}=\mathcal{F}{c_0}+\mathcal{F}{n_0}$. This fused feature map replaces the original inputs for the first Up-Blocks of the two U-Nets, $UB{c1}(\cdot)$ and $UB_{n1}(\cdot)$, leading to their respective new outputs: $\mathcal{F}{c_1}$ and $\mathcal{F}{n_1}$. This process is defined by the formulas: $\mathcal{F}{c_1}=UB{c1}(\mathcal{F}{f_0})$, $\mathcal{F}{n_1}=UB_{n1}(\mathcal{F}_{f_0})$.

We apply the same residual connections to $\mathcal{F}{c_1}$ and $\mathcal{F}{n_1}$ and repeat this series of operations from Up-Block-1s to Up-Block-2s, continuing this interactive process through to Up-Block-5s. This approach deeply integrates the two U-Nets, allowing them to interact at multiple layers, harmonizing the relationship between the cross-modal features of texture and geometry, ultimately producing more refined Gaussian avatars.


\myparagraph{Gaussian Enhanced Remeshing Strategy} Building on the generated texture and normal Gaussian avatars, we introduce our Gaussian enhanced remeshing strategy to achieve high-fidelity textured 3D human meshes for downstream applications. Previous approaches have attempted to derive human or object meshes from Gaussian representations~\cite{zhang2024multigo, 3diffusion, tang2024lgm} or normal maps~\cite{li2024pshuman, unique3d}. However, these methods often produce inaccurate results due to hallucinations and multi-view inconsistencies introduced by diffusion models during extraction or post-processing, and they can also suffer from low computational efficiency.

In contrast, our approach effectively utilizes the ``by-product" normal Gaussian avatar generated by the reconstruction network. This strategy not only addresses the challenges of multi-view inconsistency and model hallucination by leveraging the inherent multi-view consistency of 3D Gaussian representations, but also offers significantly improved computational efficiency compared to mesh extraction pipelines based on implicit functions~\cite{yu2024gaussian}.

Particularly, we begin by initializing a coarse mesh using the mesh conversion technique from~\cite{tang2024lgm} with $G^n$. Utilizing this initialized mesh, we apply differentiable rendering~\cite{nvdiffrast} to optimize the 3D geometry with $G^n$. The optimization targets consist of the normal maps and masks rendered from $G^n$. Our goal is to refine the geometry by minimizing the discrepancies between the normal map and mask rendered from the coarse mesh and their respective target counterparts. The objective loss function of the remeshing process is defined as follows:

Initially, we start by creating a coarse mesh using the mesh conversion technique described in~\cite{tang2024lgm} with $G^n$. With this initialized mesh, we employ differentiable rendering~\cite{nvdiffrast} to optimize the 3D geometry using $G^n$. The optimization process focuses on the normal maps and masks rendered from $G^n$. We aim to refine the geometry by minimizing the differences between the normal map and mask rendered from the coarse mesh and their respective target versions. The objective loss function for the remeshing process is defined as follows:

\begin{equation}
\begin{aligned}
    \mathcal{L}_{remesh} = \mathcal{L}_{normal} + \mathcal{L}_{mask} + \mathcal{R}_{Lap},
\end{aligned}
\end{equation}
where $\mathcal{L}_{normal}$ represents the $L_2$ loss between the rendered normals and the target normals, $\mathcal{L}_{mask}$ denotes the $L_2$ loss between the rendered masks and the target masks, and the $\mathcal{R}_{Lap}$ is the Laplace regularization term to control the mesh smoothness. 

\begin{table*}
    \centering
    \caption{\textbf{Comparison of Human Geometry with SOTA methods.}  The \textbf{best} and \underline{second} results are highlighted with bold and underline respectively. Arrow $\uparrow$/$\downarrow$ means higher/lower is better. \colorbox{gray!20}{Grey} background represents the test set is used as training data, and these methods are excluded from the ranking. ``$^{\dagger}$'' indicates the models trained on more commercial or synthesis data. \label{main_exp_3d}}
    \vspace{-0.1cm}
    \renewcommand{\arraystretch}{0.95}
    \scalebox{0.9}{
    \begin{tabular}{@{\extracolsep{\fill}}l|c|ccc|ccc}
    \toprule
         \multirow{2}{*}{Methods}  & \multirow{2}{*}{Publication}     & \multicolumn{3}{c}{CustomHuman~\cite{ho2023customhuman}} & \multicolumn{3}{c}{THuman3.0~\cite{thuman3.0}}  \\
                                   &                               & \begin{tabular}{c} CD: P-to-S /\\ S-to-P $(\mathrm{cm}) \downarrow$ \end{tabular} & NC $\uparrow$ & F-score $\uparrow$ &
                                                                    \begin{tabular}{c} CD: P-to-S /\\ S-to-P $(\mathrm{cm}) \downarrow$ \end{tabular} & NC $\uparrow$ & F-score $\uparrow$  \\
    \midrule 
    PIFu~\cite{saito2019pifu}         & ICCV 2019   & $2.965/3.108$ & 0.765 & 25.708  & $2.176/2.452$   & 0.773 & 34.194 \\
    ICON~\cite{xiu2022icon}          & CVPR 2022   & $2.441/2.823$ & 0.785 & 29.144 & $2.368/2.776$   & 0.754 & 27.434 \\
    ECON~\cite{xiu2023econ}          & CVPR 2023   & $2.196/2.340$ & 0.801 & 33.292 & $2.201/2.271$   & 0.783 & 33.223 \\   
    GTA~\cite{zhang2023global}       & NeurIPS 2023    & $2.404/2.726$ & 0.790 & 29.907  & $2.416/2.652$   & 0.768 & 29.257 \\   
    
    VS~\cite{VS_CVPR2024}            & CVPR 2024   & $2.518/2.993$ & 0.780 & 26.791 & $2.526/2.942$   & 0.753 & 26.344 \\    
    HiLo~\cite{hilo}                 & CVPR 2024   & $2.282/2.741$ & 0.792 & 30.282 & $2.395/2.872$   & 0.770 & 28.120 \\
    SIFU~\cite{Zhang_2024_sifu}      & CVPR 2024   & $2.460/2.780$ & 0.784 & 28.564 & $2.450/2.832$   & 0.772 & 27.921 \\
    SiTH~\cite{ho2024sith}           & CVPR 2024   & $1.832/2.148$ & 0.826 & 36.154 & $1.743/2.019$   & 0.774 & 36.274 \\
    HumanRef~\cite{zhang2024humanref}& CVPR 2024   & $2.073/2.228$ & 0.812 & 34.469 & $1.975/2.248$   & 0.783 & 34.506 \\
    FOF-X~\cite{feng2024fof}         & \red{TMM 2026}   & $1.725/1.951$ & 0.823 & 39.794 & $	1.681/1.826$   & 0.813 &  39.872 \\
    R$^2$Human~\cite{yang2024r}         & ISMAR 2024   & $2.129/2.366$ & 0.799 & 32.185 & $	2.123/2.332$   & 0.775 &  31.314 \\
    \rowcolor{gray!20}
    H3Diff.$^{\dagger}$~\cite{3diffusion}  & NeurIPS 2024& $1.481/1.505$ & 0.864 & 47.019 & $1.331/1.456$   &   0.843   &   49.639   \\
    \rowcolor{gray!20}
    PSHuman~\cite{li2024pshuman}                  & CVPR 2025   & $1.923/2.046$  & 0.830   & 36.899     & $1.827/1.844$   &   0.796   &   38.855      \\
    MultiGO~\cite{zhang2024multigo}               & CVPR 2025   & $1.620/1.782$  & 0.850   & 42.425     & $1.408/1.633$   &   0.834   &   46.091   \\
    \midrule
    \projecttitle               &  -  &   $\underline{1.482}/\underline{1.652}$  &  \underline{0.859}  &   \underline{45.038} &   $\underline{1.237}/\underline{1.406}$ &  \underline{0.842}&  \underline{51.012}    \\ 
    
    \projecttitle$^{\dagger}$   &  -  &   $\textbf{1.402}/\textbf{1.562}$  &  \textbf{0.865}  &   \textbf{47.208} &  $\textbf{1.173}/\textbf{1.299}$ &  \textbf{0.850}   &  \textbf{53.480}    \\ 

    \bottomrule 
    \end{tabular} 
    } 
    \vspace{-0.2cm}
\end{table*}

\begin{table*}[t!]
    \centering
    \caption{\textbf{Comparison of Human Texture with SOTA methods.} For comparison, we render the textured 3D human reconstruction results of these methods in front view and back view, represented by ``F/B" symbols. Note that only partial methods reconstruct 3D human texture, and ICON and ECON only reconstruct the front-view texture. 
    \label{main_exp_2d}}
    \renewcommand{\arraystretch}{0.95}
    \vspace{-0.1cm}
    \scalebox{0.9}{
    \begin{tabular}{@{\extracolsep{\fill}}l|ccc|ccc}
    \toprule
         \multirow{2}{*}{Methods}  & \multicolumn{3}{c}{CustomHuman} & \multicolumn{3}{c}{THuman3.0}  \\
                                   & LPIPS: F/B $\downarrow$ & SSIM: F/B $\uparrow$ & PSNR: F/B $\uparrow$ &
                                     LPIPS: F/B $\downarrow$ & SSIM: F/B $\uparrow$ & PSNR: F/B $\uparrow$   \\
    \midrule
    PIFu~\red{\cite{saito2019pifu}}    & $0.0792/0.0966$ & $0.8965/0.8742$ & $18.141/16.721$ & $0.0706/0.0849$ & $0.9242/0.9007$ & $20.104/17.926$\\  
    ICON~\red{\cite{xiu2022icon}}    & $0.0714/-$ & $0.8975/-$ & $18.614/-$ & $0.0602/-$ & $0.9287/-$ & $21.126/-$\\     
    ECON~\red{\cite{xiu2023econ}}    & $0.0777/-$ & $0.8870/-$ & $18.437/-$ & $0.0638/-$ & $0.9258/-$ & $20.951/-$ \\ 
    GTA~\red{\cite{zhang2024global_gta}}     & $0.0730/0.0891$ & $0.9003/0.8923$ & $18.790/18.229 $ & $0.0633/0.0770$ & $0.9298/0.9275$ & $21.113/20.497$ \\ 
    
    SIFU~\red{\cite{Zhang_2024_sifu}}    & $0.0682/0.0880$ & $0.9018/0.8907$ & $18.710/18.114$ & $0.0594/0.0764$ & $0.9307/0.9245$ & $21.103/20.351$ \\
    SiTH~\red{\cite{ho2024sith}}    & $0.0667/0.0841$ & $0.9010/0.8873$ & $18.420/17.613$ & $0.0618/0.0770$ & $0.9233/0.9110$ & $20.324/19.353$ \\
    R$^2$Human~\red{\cite{yang2024r}}    & $0.0776/0.0908$ & $0.8974/0.8844$ & $18.485/17.460$ & $0.0654/0.0829$ & $0.9280/0.9126$ & $20.598/19.226$ \\ 
    HumanRef~\red{\cite{zhang2024humanref}}    & $0.0686/0.0862$ & $0.9101/0.8995$ & $19.592/18.902$ & $0.0603/0.0767$ & $0.9373/0.9280$ & $21.783/19.942$ \\ 
    FOF-X~\red{\cite{feng2024fof}}    & $0.0628/0.0746$ & $0.9384/0.9169$ & $21.905/19.865$ & $0.0578/0.0692$ & $0.9473/0.9381$ & $22.730/21.587$ \\ 
    \rowcolor{gray!20}
    H3Diff.$^{\dagger}$~\red{\cite{3diffusion}} & $0.0569/0.0641$ & $0.9398/0.9352$ & $20.909/20.436$ & $0.0540/0.0610$ & $0.9553/0.9498$ & $23.402/22.032$ \\
    \rowcolor{gray!20}
    PSHuman~\red{\cite{li2024pshuman}} & $0.0647/0.0717$ & $0.9069/0.9024$ & $18.859/18.564$ & $0.0587/0.0641$ & $0.9302/0.9338$ & $21.165/21.137$ \\
    MultiGO~\red{\cite{zhang2024multigo}}  & $0.0414/0.0643$ & $0.9603/0.9415$ & $22.347/20.849$ & $0.0457/0.0616$ & $0.9623/0.9512$ & $23.794/22.657$ \\    
    \midrule
    \projecttitle  & 
    $\underline{0.0411}/\underline{0.0641}$ & $\underline{0.9654}/\underline{0.9429}$ & $\underline{23.122}/\underline{20.865}$ & $\underline{0.0377}/\underline{0.0588}$ & $\underline{0.9826}/\underline{0.9630}$ & $\underline{26.358}/\underline{23.920}$ \\    
    \projecttitle$^{\dagger}$  & 
    $\textbf{0.0372}/\textbf{0.0599}$ & 
    $\textbf{0.9679}/\textbf{0.9475}$ &
    $\textbf{23.646}/\textbf{21.313}$ &
    $\textbf{0.0368}/\textbf{0.0567}$ & 
    $\textbf{0.9842}/\textbf{0.9646}$ &
    $\textbf{26.747}/\textbf{24.201}$ \\    

    \bottomrule
    \end{tabular}
    } 
    \vspace{-0.5cm}
\end{table*}

\section{Experiment}\label{sec: exp}
\subsection{Experiment Setup}
\noindent \textbf{Datasets.} Our basic model is trained using the widely recognized 3D human scan dataset, THuman 2.0~\cite{tao2021function4d_thuman}. For evaluation purposes, we utilize the CustomHuman benchmark~\cite{ho2023customhuman} and the THuman 3.0 benchmark~\cite{thuman3.0}, as introduced by SiTH~\cite{ho2024sith} and MultiGO~\cite{zhang2024multigo}, respectively. To ensure fair comparisons, we optionally integrate both commercial and synthesized human scans into our training data. Importantly, our training method does not depend on additional annotated SMPL-related parameters. \red{For detailed information regarding the synthetic and commercial datasets employed, readers are directed to the Supplementary Material.}

\noindent \textbf{Training \& Inference} We conducted our experiments on a server equipped with eight NVIDIA A800 GPUs. Leveraging the well-established research in this area, we employed a fine-tuning strategy for training our models. During the training stage, we set the batch size to 1 and utilized the AdamW~\cite{adamw} optimizer with a learning rate of $1\times10^{-5}$. We used 8-view orthographic RGB and normal maps, rendered from 3D scans, as supervision for model training. Our loss functions included MSE loss, LPIPS loss, and mask loss, with the LPIPS loss weighted at 2 and the others at 1. The LPIPS loss was calculated using the VGG-16 model. \red{The training process takes approximately 72 GPU hours for the model to converge.} In the inference stage, all input images were rendered at a resolution of 512 $\times$ 512 using Nvdiffrast~\cite{Laine2020diffrast}, and backgrounds were removed using Rembg to ensure a fair comparison. For our method, the rendered low-resolution images were then upsampled to 896 $\times$ 896 to meet the input requirements of the Vision Transformer.


\noindent \textbf{Evaluation Metrics.} In line with prior research~\cite{ho2024sith,zhang2024multigo}, we utilize three 3D metrics for assessing the geometric accuracy of our generated meshes: Chamfer Distance (\textbf{CD}), Normal Consistency (\textbf{NC}), and \textbf{F-score}~\cite{fscore}. For evaluating texture quality, we calculate the Peak Signal-to-Noise Ratio (\textbf{PSNR}), Structural Similarity Index Measure (\textbf{SSIM}), and Learned Perceptual Image Patch Similarity (\textbf{LPIPS})~\cite{lpips} on both front and back views. Moreover, to assess the computational efficiency of our approach, we measure the inference time (\textbf{Infer. Time}) for each method. Specifically, for the Gaussian-based method, we also account for the time overhead incurred during the mesh extraction process (\textbf{M.E. Time}).

\subsection{Evaluation}

\myparagraph{Quantitative Evaluation on Geometry} Table~\ref{main_exp_3d} highlights the notable performance of our proposed MultiGO++ on the CustomHuman and THuman3.0 benchmarks regarding reconstructed geometry quality. Our approach consistently outperforms SOTA methods, including those based on implicit functions~\cite{xiu2022icon, xiu2023econ, VS_CVPR2024, hilo, ho2024sith, Zhang_2024_sifu, feng2024fof, zhang2024humanref, yang2024r}, Gaussian models~\cite{zhang2024multigo, 3diffusion}, and diffusion techniques~\cite{li2024pshuman}. Specifically, compared to the leading existing method, MultiGO~\cite{zhang2024multigo}, MultiGO++ achieves improvements of 0.218/0.220 in CD, 0.015 in NC, and 4.783 in F-score on CustomHuman. For the THuman3.0 benchmark, it achieves enhancements of 0.235/0.334 in CD, 0.016 in NC, and 7.389 in F-score. Remarkably, even under test data leak conditions, Human3Diffusion and PSHuman fail to match the performance of MultiGO++ on THuman3.0. These findings underscore the effectiveness and robustness of MultiGO++ in accurately reconstructing human geometry across various challenging scenarios.

\begin{figure*}
    \centering
    \includegraphics[width=1\linewidth]{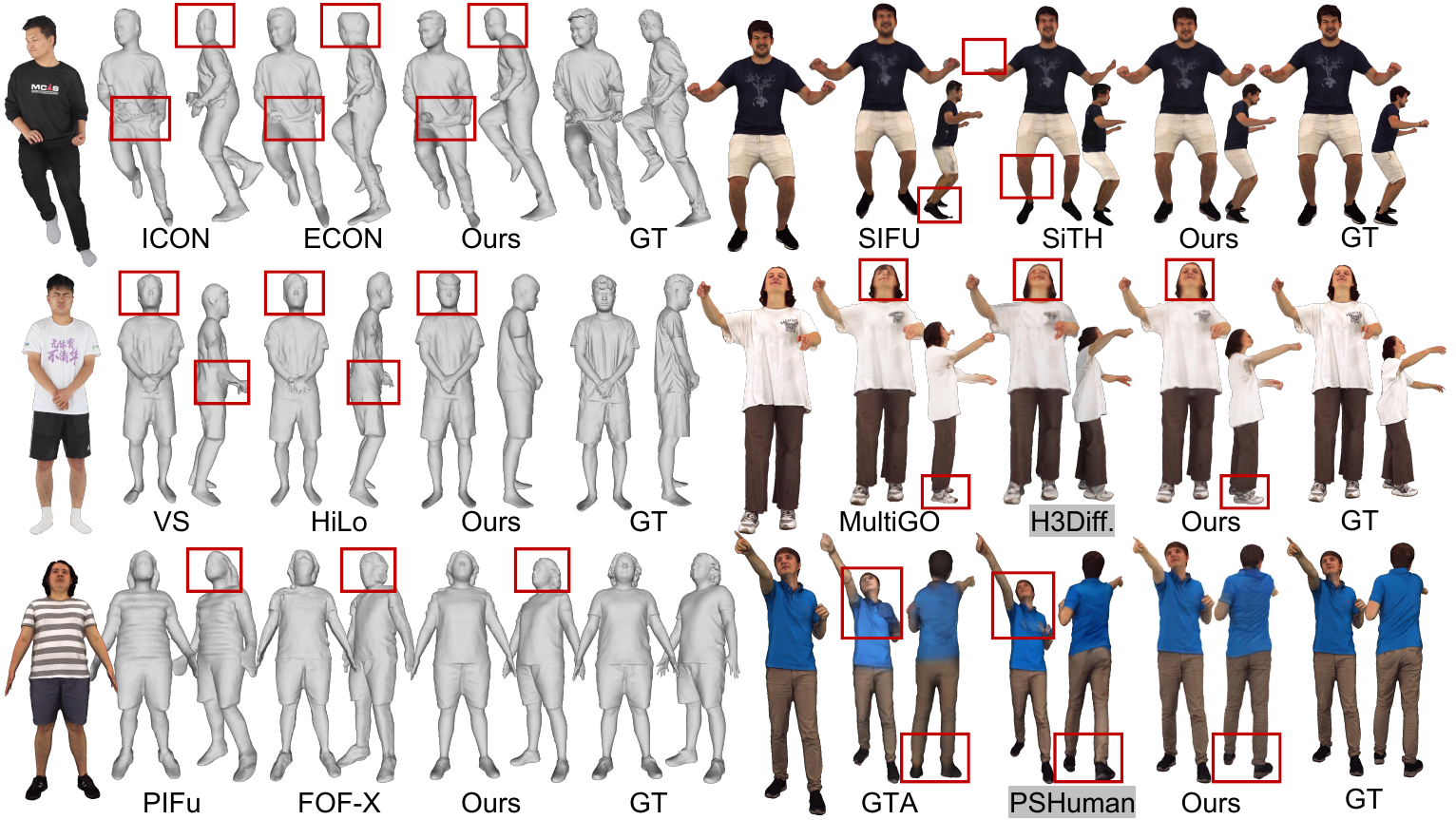}
    \vspace{-0.6cm}
    \caption{\textbf{Visual Comparisons with SOTA methods.} Previous SOTA approaches always struggle with recovering the correct human pose, shape, and fine-grained geometry and texture details. For a detailed view, please 
    \textbf{zoom in}~\includegraphics[width=0.015\textwidth]{fig/icon/fangdajing.png}~on the image.} 
    \label{fig: viscmp}
    \vspace{-0.5cm}
\end{figure*}

\myparagraph{Quantitative Evaluation on Texture Quality} 
The reconstructed texture quality, as detailed in Table~\ref{main_exp_2d}, also highlights the clear advantage of \projecttitle~over existing SOTA methods. Concretely, \projecttitle~achieves notable gains LPIPS by 0.0042/0.0044 (F/B), SSIM by 0.0076/0.0060 (F/B), and PSNR by 1.299/0.464 (F/B) on CustomHuman and LPIPS by 0.0089/0.0049 (F/B), SSIM by 0.0219/0.0134 (F/B), and PSNR by 2.953/1.634 on THuman3.0, respectively. These findings underscore the robustness of \projecttitle~in generating high-fidelity textured 3D avatars compared to other approaches. 

\begin{figure*}[t!]
    \centering
    \includegraphics[width=1.0\linewidth]{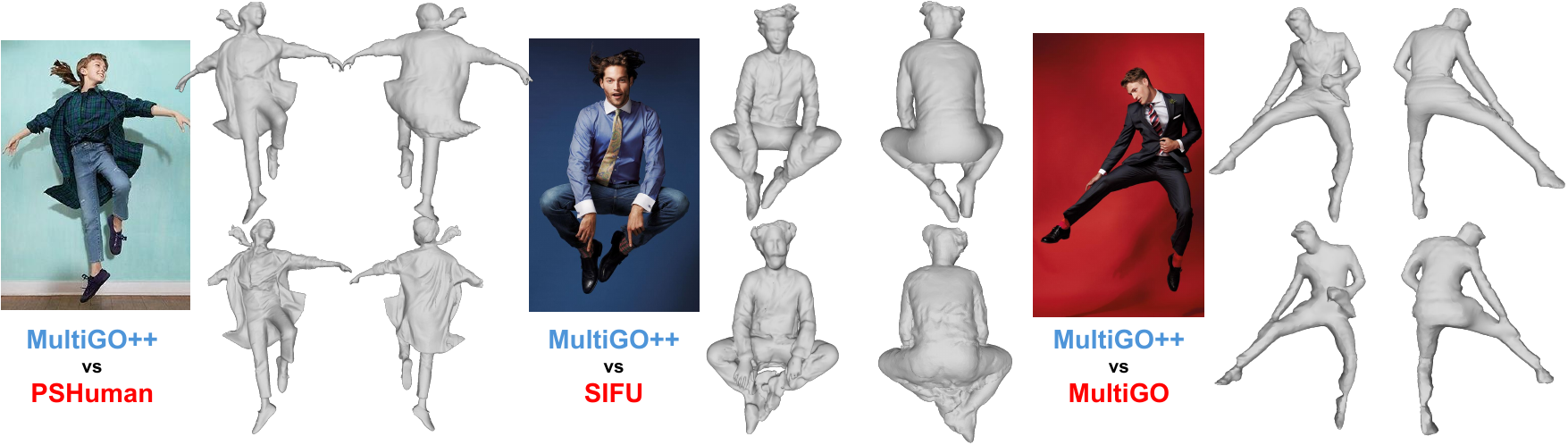}
    \vspace{-0.7cm}
    \caption{\red{\textbf{Qualitative comparisons on in-the-wild images featuring challenging poses.} Compared to existing approaches, our method robustly reconstructs geometrically accurate limb structures and preserves fine-grained facial expressions, while introducing significantly fewer visual artifacts.} Please \textbf{zoom in}~\includegraphics[width=0.015\textwidth]{fig/icon/fangdajing.png}~for a detailed view.}
    \label{fig: viswild-poses}
    \vspace{-0.5cm}
\end{figure*}

\myparagraph{Qualitative Evaluation} The outcomes of the visual comparison are illustrated in Fig.~\ref{fig: viscmp}. Both the ICON and ECON methods exhibit notable shortcomings in accurately reconstructing intricate features of the hands and head. HiLo and VS display less-than-ideal performance, particularly when faced with complex finger arrangements. SIFU has difficulty in maintaining accurate human poses, while SiTH struggles with incomplete reconstructions of the hands. Additionally, MultiGO and Human3Diffusion are unable to effectively recover facial textures, especially in non-frontal views. \red{PIFu is limited in its reconstruction fidelity due to the absence of explicit pose priors. While FOF-X employs Fourier feature encodings similar to our method, it remains sensitive to pose estimation errors during inference, leading to unsatisfactory reconstruction quality. Furthermore, GTA and PSHuman struggle to resolve geometric details when processing inputs with severe depth ambiguities. To further assess the capabilities of \projecttitle~in managing complex scenarios like loose clothing and challenging poses, we conducted experiments and comparisons on in-the-wild images, as depicted in Fig.~\ref{fig: viswild} and Fig.~\ref{fig: viswild-poses}. These findings underscore the robust generalization ability of \projecttitle~under complex conditions. For additional visualizations and evaluations, please refer to the Supplementary Material.}

\begin{table}
\centering
 \caption{\textbf{Evaluation on Computational Efficiency.} Our MultiGO++ achieves the fastest mesh extraction while maintaining highly competitive inference time.}
 \label{tab:speed}
 \scalebox{0.75}
	{
    \begin{tabular}{l|ccccc}
    \toprule
   Method       & ICON & ECON   & GTA    & VS     & HiLo  \\
   \midrule
   Infer. Time  & 20s  & 8.5min & 4.5min & 20min  & 1.5mins \\
   \midrule
   Method       & SIFU & SiTH   & R$^2$Human & HumanRef & PSHuman   \\
   \midrule
   Infer. Time  & 6mins & 2mins & 2s      & 2h        & 50s \\
   \midrule
   Method       & H3Diff. & MultiGO & MultiGO++ & ~  \\
   \midrule
   Infer. Time  & 2mins & \textbf{0.6s}   & \underline{0.7s} & ~  & ~ \\
   M.E. Time    & 12min & \underline{3min} & \textbf{1min} & ~  & ~ \\
    \bottomrule
    \end{tabular}
	}
    \vspace{-0.6cm}
\end{table}


\myparagraph{Evaluation on Computational Efficiency} Table~\ref{tab:speed} presents the statistical results for computational efficiency across various methods. Our proposed approach, MultiGO++, demonstrates exceptional performance in both metrics. It boasts a remarkably swift inference time of just 0.7 seconds, significantly outperforming most other methods. For example, approaches like ECON, GTA, and SIFU have inference times of 8.5 minutes, 4.5 minutes, and 6 minutes, respectively, making MultiGO++ considerably faster. Even recent methods based on Gaussian diffusion, such as Human3Diffusion, require 2 minutes for inference. Although MultiGO has half the reconstruction backbone network parameters of MultiGO++, its efficiency is hampered by the optimization-based HPS method. Nevertheless, MultiGO++ achieves comparable inference speed, highlighting the impressive efficiency of its core inference process. Furthermore, a notable computational challenge in Gaussian-based methods is the subsequent mesh extraction stage. MultiGO++ significantly enhances this aspect, reducing mesh extraction time to just 1 minute—a threefold improvement over MultiGO (3 minutes) and a twelvefold improvement over Human3Diffusion (12 minutes). This advancement is crucial for enhancing overall pipeline efficiency, ensuring that our method is not only rapid in generating initial results but also highly effective in delivering the final high-quality 3D mesh output.

\begin{figure}
    \centering
    \includegraphics[width=0.75\linewidth]{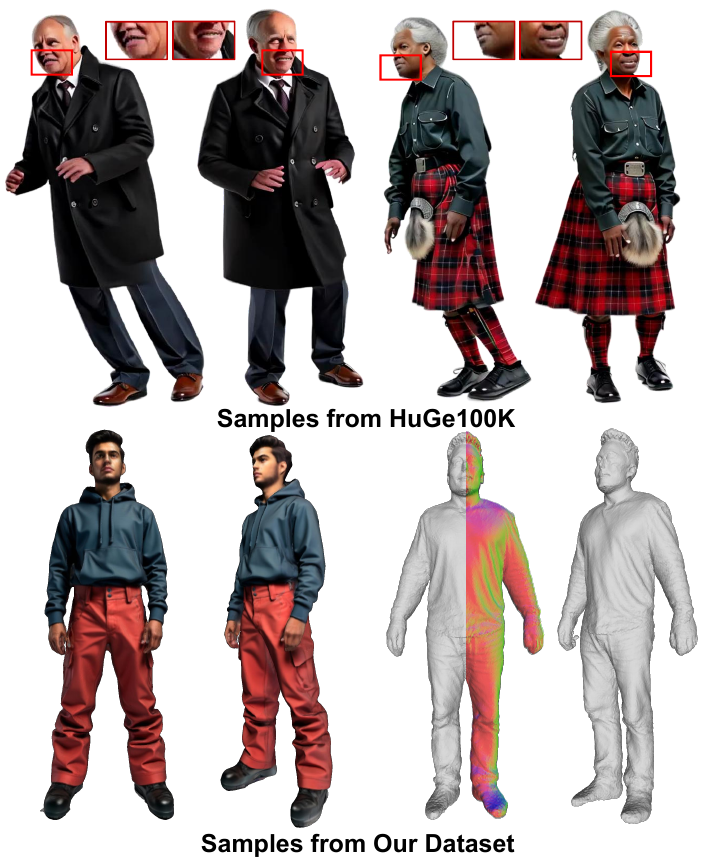}
    \vspace{-0.3cm}
    \caption{\textbf{Quality Evaluation on Synthetic data.} We show randomly selected data from HuGe-100K and our dataset, where locations with cross-view inconsistencies are marked with red boxes. Please \textbf{zoom in}~\includegraphics[width=0.015\textwidth]{fig/icon/fangdajing.png}~for a detailed view.}
    \label{fig: datacmp}
    \vspace{-0.6cm}
\end{figure}


\myparagraph{Evaluation of Synthetic Dataset Quality} To assess the quality of our synthetic data, we compare it with the publicly available 3D synthetic human dataset, HuGe-100K~\cite{zhuang2024idolinstantphotorealistic3d}. HuGe-100K is a synthetic video human dataset created using a modified Image-to-Video generation model~\cite{zhu2024champ}. As illustrated in Fig.~\ref{fig: datacmp}, the synthetic data in HuGe-100K suffers from limitations inherent to the video generation model, resulting in notable inconsistencies across different viewpoints, particularly in intricate details like facial expressions. In contrast, our method employs a mesh-based representation that ensures strict cross-view consistency during rendering. Furthermore, the explicit and continuous surface topology of the chosen mesh representation allows for the rendering of high-quality normal maps, capturing fine geometric details. This advancement enhances the dataset's applicability across a broader spectrum of methods.


\begin{table*}[!t]
    \centering
    \renewcommand{\arraystretch}{0.95}
    \caption{\textbf{Ablation Study on Geometry Accuracy.}\label{tbl: abl_exp_3d} \textit{Texture}: To evaluate the effectiveness of texture synthesis strategy proposed in section on model performance by comparing results using synthetic data against high-quality commercial training data. \textit{Geometry}: we conduct an ablation study by \red{replacing} the Region-aware Shape Extraction Module (RSEM) during inference and evaluate the effectiveness of the Fourier Geometry Encoder (FGE) for reconstructing human geometry by comparing model performance with and without the 2D projection of 3D features. \textit{System}: We assess the contributions of the dual U-Net by separately ablated normal U-Net and the Gaussian enhanced remeshing strategy. }
    \scalebox{0.9}{
    \begin{tabular}{@{\extracolsep{\fill}} l|c|ccc|ccc}
    \toprule
         \multirow{3}{*}{Methods} & \multirow{3}{*}{Section} & \multicolumn{3}{c}{CustomHuman~\cite{ho2023customhuman}} & \multicolumn{3}{c}{THuman3.0~\cite{thuman3.0}}  \\
                  & & \begin{tabular}{c} CD: P-to-S /\\ S-to-P $(\mathrm{cm}) \downarrow$ \end{tabular} & NC $\uparrow$ & F-score $\uparrow$ &
                                                                    \begin{tabular}{c} CD: P-to-S /\\ S-to-P $(\mathrm{cm}) \downarrow$ \end{tabular} & NC $\uparrow$ & F-score $\uparrow$  \\
    \midrule 
    w/  $Data_{Com.+Syn.}$    & \multirow{3}{*}{Texture}       & $1.402/1.562$ & 0.865 & 47.208 & $1.173/1.299$   & 0.850 & 53.480 \\
    w/  $Data_{Com.}$         &                                 & $1.476/1.632$ & 0.860 & 45.972 & $1.214/1.318$   & 0.840 & 52.545 \\
    w/o $Data.$               &                                 & $1.481/1.652$ & 0.859 & 45.038 & $1.237/1.406$   & 0.842 & 51.012 \\
    \midrule
    w/ 3-view Proj.    &\multirow{5}{*}{Geometry} & $1.462/1.658$ & 0.858 & 45.081 & $1.221/1.348$ & 0.841 & 52.128 \\ 
    w/ 2-view Proj.    &                         & $1.546/1.761$ & 0.845 & 42.884 & $1.267/1.831$ & 0.834 & 50.621 \\
    w/ 1-view Proj.    &                         & $1.771/1.913$ & 0.823 & 43.635 & $1.342/2.040$ & 0.819 & 49.883 \\  
    w/o FGE  &                         & $2.147/2.416$ & 0.823 & 39.624 & $1.666/2.160$ & 0.822 & 48.086 \\
    
   \red{w/ Simplify} &                 & $1.580/1.726$ & 0.843 & 43.673 & $1.408/1.534$ &  0.823 & 48.005 \\
    \red{w/ HMR2.0}   &                 & \red{$1.513/1.667$} & \red{0.849} & \red{44.512} & \red{$1.300/1.423$} &  \red{0.838} & 
    \red{49.512} \\
    
    \midrule

    w/o Remeshing                                                     & \multirow{3}{*}{System}  & $1.489/1.609$ & 0.863 & 45.416 &  $1.259/1.397$& 0.847 & 51.039 \\   
    \begin{tabular}{l} w/o Normal U-Net \\ \&  Remeshing \end{tabular}&  & $1.518/1.625$ & 0.859 & 45.102 & $1.284/1.431$ & 0.846 & 50.793 \\    
    
    \midrule
    \projecttitle  & - & $1.402/1.562$ & 0.865 & 47.208 & $1.173/1.299$   & 0.850 & 53.480 \\
    \bottomrule 
    \end{tabular} 
    } 
    \vspace{-0.40cm}
\end{table*}

\begin{figure}
\begin{minipage}[!t]{0.48\textwidth}
  \centering
    \captionof{table}{\textbf{Ablation Study on Texture Quality.} We demonstrate the reconstructed texture quality is also improved with our proposed region-aware shape extraction module, Fourier geometry encoder and multi-source texture synthesis strategy.\label{tbl: 2dabl}}
    \scalebox{0.70}{
    \begin{tabular}{l|ccc}
    \toprule
    CustomHuman & LPIPS: F/B $\downarrow$ & SSIM: F/B $\uparrow$ & PSNR: F/B $\uparrow$  \\
    \midrule
        w/ $Data_{Com.+Syn.}$ & $0.0372/0.0599$ & $0.9679/0.9475$ & $23.646/21.313$ \\
        w/ $Data_{Com.}$  & $0.0407/0.0634$ & $0.9666/0.9451$ & $23.412/21.012$ \\
        w/o $Data$        & $0.0414/0.0641$ & $0.9654/0.9423$ & $23.122/20.865$ \\
    \midrule
        \red{w/ Simplify}              & $0.0443/0.0667$ & $0.9588/0.9364$ & $22.569/20.456$ \\
        \red{w/ HMR2.0} & \red{$0.0412/0.0610$} & \red{$0.9621/0.9422$} & \red{$22.912/20.707$} \\
        w/o FGE                 & $0.0462/0.0689$ & $0.9502/0.9312$ & $22.437/20.347$ \\
    \midrule
        \projecttitle        & $0.0372/0.0599$ & $0.9679/0.9475$ & $23.646/21.313$ \\
    \midrule
    THuman3.0 & LPIPS: F/B $\downarrow$ & SSIM: F/B $\uparrow$ & PSNR: F/B $\uparrow$   \\
    \midrule
        w/ $Data_{Com.+Syn.}$ & $0.0368/0.0567$ & $0.9842/0.9646$ & $26.747/24.201$ \\
        w/ $Data_{Com.}$ & $0.0374/0.0586$ & $0.9828/0.9633$ & $26.377/23.907$ \\
        w/o $Data$       & $0.0377/0.0588$ & $0.9826/0.9632$ & $26.358/23.920$ \\ 
    \midrule
        \red{w/ Simplify} & $0.0416/0.0612$ & $0.9733/0.9554$   & $25.922/22.994$ \\  
        \red{w/ HMR2.0} & \red{$0.0395/0.0580$} & \red{$0.9800/0.9611$} & \red{$26.133/23.802$} \\
        w/o FGE & $0.0452/0.0635$ & $0.9702/0.9501$ & $25.442/22.431$ \\
    \midrule
        \projecttitle    & $0.0368/0.0567$ & $0.9842/0.9646$ & $26.747/24.201$ \\    
    \bottomrule
    \end{tabular}
    }
\label{tab: abl_2d}
\end{minipage}
\vspace{-0.5cm}
\end{figure}


\subsection{Ablation Study}

\myparagraph{Effectiveness of Synthetic Texture} Textually, the results presented in Tables~\ref{tbl: 2dabl} and ~\ref{tbl: abl_exp_3d} underscore the efficacy of our data synthesis approach, particularly in enhancing texture reconstruction performance. We evaluated our complete model (w/ $Data_{Com.+Syn.}$) against two alternative configurations: one that utilized only high-quality commercial data (w/ $Data_{Com.}$) and another that did not incorporate any additional texture data (w/o $Data.$). The findings consistently demonstrate a performance gradient across all metrics on both benchmarks. This progressive improvement reinforces the notion that our synthetic texture strategy significantly enhances the training data. It offers complementary texture and geometric information that the model utilizes to achieve more accurate texture and shape estimations, outperforming what can be achieved with high-quality commercial data alone. This highlights the importance of our multi-source texture synthesis strategy in attaining high-fidelity reconstruction.

\myparagraph{Effectiveness of the Shape Extraction Module and Fourier Geometry Encoder} Tables~\ref{tbl: 2dabl} and ~\ref{tbl: abl_exp_3d} illustrate the benefits of our proposed region-aware shape extraction module and Fourier geometry encoder. \red{In the ``w/ Simplify'' and ``w/ HMR2.0'' setting, we replace the region-aware shape extraction module with a widely recognized pose estimation method~\cite{smplify} and current SOTA method~\cite{Goel_2023_ICCV}}. In the ``w/o FGE" scenario, we encode 3D geometry Fourier features as a whole using multiple convolutional layers, rather than first projecting the 3D geometry Fourier data into 2D features as we have proposed. The ``1-view Proj.,'' ``2-view Proj.,'' and ``3-view Proj.'' settings implement our projection operation using one, two, and three camera views, respectively. The quantitative results highlight the essential roles of both components. \red{The removal of the region-aware shape extraction module ``w/ Simplify'' consistently results in performance degradation across both datasets, indicating that this region-aware approach effectively captures accurate human body pose and shape, which leads to improved shape reconstruction. Crucially, our method outperforms the setting utilizing the SOTA estimator ``w/ HMR2.0''. This suggests that simply relying on global parametric regression is insufficient for reconstruction tasks requiring fine-grained geometry. In contrast, our RSEM leverages cross-attention to facilitate interaction among local body regions, thereby effectively mitigating depth ambiguity and ensuring better feature alignment than external pose priors.} Omitting the Fourier geometry encoder (w/o FGE) results in the most substantial decline in performance across all metrics, reinforcing that our 2D projection strategy is crucial for effectively encoding 3D geometric information. Additionally, we observe a strong positive correlation between the number of projection views and reconstruction accuracy, with performance steadily improving as the number of views increases from one to three. The full model achieves the best results, highlighting the significance and effectiveness of 2D-3D modality fusion for comprehensive geometric learning.

\begin{figure*}
    \centering
    \includegraphics[width=0.95\linewidth]{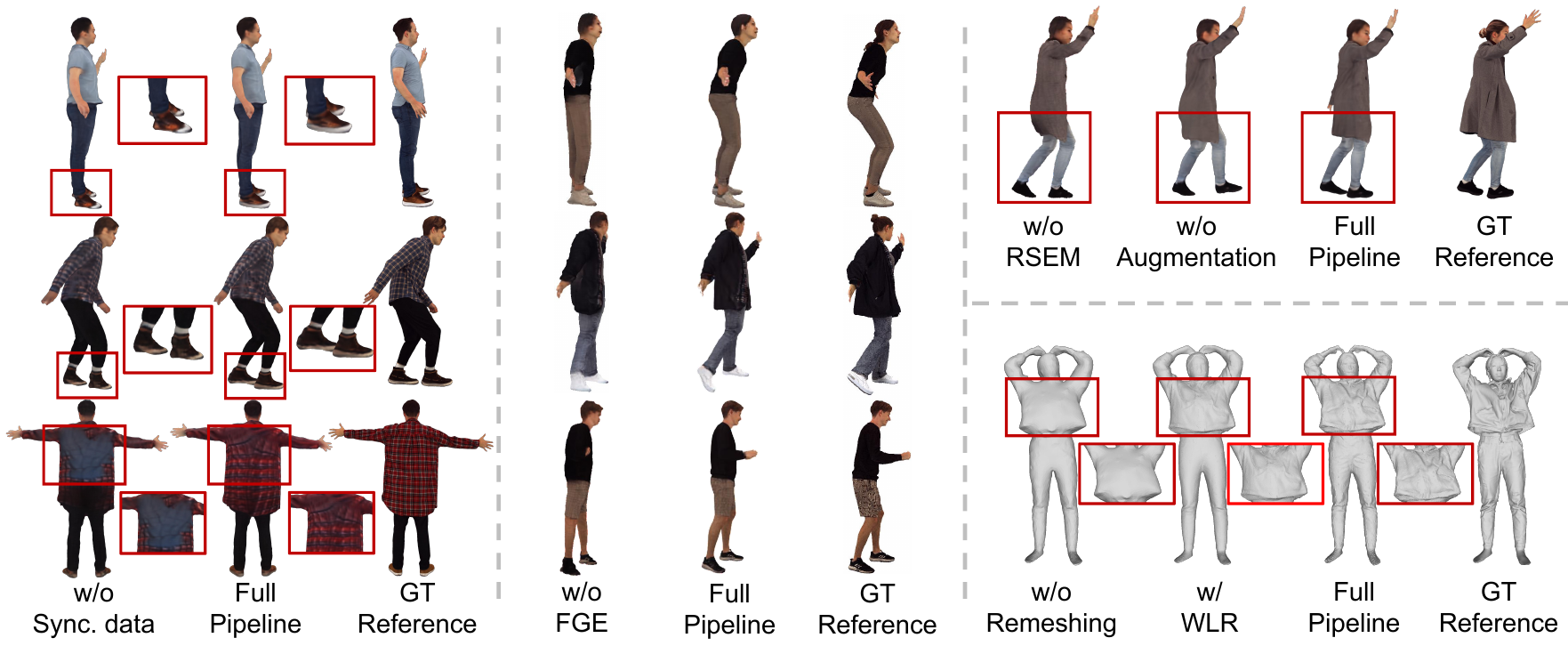}
    \vspace{-0.2cm}
    \caption{\textbf{Visual Ablation.} (\textbf{\textit{left}}) Synthetic data improves the model performance on human texture estimation. (\textbf{\textit{middle}}) The Fourier geometry encoder enhances the better feature fusion between 2D and 3D, enabling the overall human poses close to the ground truth. (\textbf{\textit{right}}) The proposed region-aware shape extraction module improves the pose correctness. The remeshing process improves the geometry details in extracted meshes. Please \textbf{zoom in}~\includegraphics[width=0.015\textwidth]{fig/icon/fangdajing.png} to observe details.}
    \label{fig: visabl}
    \vspace{-0.5cm}
\end{figure*}

\myparagraph{Effectiveness of Normal U-Net \& Remeshing Strategy} Systematically, the results in Table~\ref{tbl: abl_exp_3d} demonstrate the individual contributions of our core components. First, excluding the geometry remeshing process leads to a significant drop in reconstruction quality, as the mesh extracted from 3DGS lacks fine geometric detail. Second, even without remeshing, the normal U-Net still provides a baseline improvement, indicating that the dual-modality supervision itself is beneficial. These findings collectively validate the effectiveness and superiority of our proposed Gaussian enhanced remeshing strategy combined with the dual reconstruction U-Net architecture.

\myparagraph{Visual Ablation}
Fig.~\ref{fig: visabl} presents an ablation study evaluating the contribution of each proposed component. In the \textit{left} subfigure, the texture synthesis strategy is shown to enhance reconstruction quality by mitigating the model's generalization limitations—noticeably for footwear (first two rows) and certain garment types (third row). The \textit{middle} subfigure demonstrates that the Fourier geometry encoder facilitates effective 2D-3D feature fusion, yielding reconstructed geometries that align more closely with the ground truth. The \textit{right} subfigure illustrates two additional improvements: In the setting of ``w/o RSEM'', we ablate the region-aware shape extraction module and replace it with the common used estimation method~\cite{smplify}. It shows that the region-aware shape extraction module increases pose correctness, especially under depth ambiguity. Additionally, we train the reconstruction model using annotated body meshes and use a region-aware shape extraction module at inference stage, as illustrated in the setting of ``w/o Augmentation''. We can also observe a significant improvement in accuracy when inputting inaccurate poses. The second row highlights how the remeshing strategy better captures fine-grained details such as clothing wrinkles and facial expressions. We further compare our approach with an alternative setup where the normal U-Net is ablated and replaced with the wrinkle-level refinement module from previous work~\cite{zhang2024multigo}. As shown in the setting of ``w/ WLR'', the baseline suffers from multi-view inconsistency, which leads to loss of detail. In contrast, our method leverages the multi-view consistency of 3DGS to produce higher-fidelity 3D human meshes for better downstream applications.

\section{Conclusion}
\label{sec: conclusion}
This paper presents MultiGO++, a comprehensive framework for high-fidelity monocular 3D clothed human reconstruction that effectively addresses the challenges of geometric inaccuracy, texture scarcity, and systematic bias inherent in prior methods. By introducing a synergistic collaboration between geometry and texture through three core innovations—a multi-source texture synthesis strategy for enhanced texture diversity, a region-aware shape extraction module coupled with a Fourier geometry encoder for robust geometric learning, and a dual reconstruction U-Net for balanced cross-modal feature and mesh refinement. Extensive evaluations on standard benchmarks and in-the-wild cases demonstrate that MultiGO++ not only surpasses existing state-of-the-art methods in accuracy and visual fidelity but also achieves significant improvements in computational efficiency. The framework's strong generalization ability to challenging real-world scenarios underscores its practicality and potential for broad applications.



\small
\bibliography{main}

\end{document}